\def\year{2022}\relax
\documentclass[letterpaper]{article} 
\usepackage{aaai22}  
\usepackage{times}  
\usepackage{helvet}  
\usepackage{courier}  
\usepackage[hyphens]{url}  
\usepackage{graphicx} 
\usepackage{natbib}  
\usepackage{caption} 
\DeclareCaptionStyle{ruled}{labelfont=normalfont,labelsep=colon,strut=off} 
\usepackage{algorithm}
\usepackage{algorithmic}
\usepackage[normalem]{ulem}
\useunder{\uline}{\ul}{}
\usepackage{xcolor}
\usepackage{amsmath}
\usepackage{amsthm}
\usepackage{amsfonts}
\usepackage[british]{babel}
\usepackage[capitalise]{cleveref}
\newcommand{\SumNode}{\mathsf{S}}
\newcommand{\ProductNode}{\mathsf{P}}
\newcommand{\Node}{\mathsf{N}}

\newcommand{\Leaf}{\mathsf{L}}
\newcommand{\mbf}[1]{\mathbf{#1}}
\usepackage{xspace}
\newcommand{\eg}{\textit{e.g.}\xspace}
\newcommand{\ie}{\textit{i.e.}\xspace}
\usepackage{newfloat}
\usepackage{listings}
\newtheorem{defin}{Definition}
\floatstyle{ruled}
\newfloat{listing}{tb}{lst}{}
\floatname{listing}{Listing}

\title{Sum-Product-Attention Networks:\\ Leveraging Self-Attention in Probabilistic Circuits}
\author{
    Zhongjie Yu,\textsuperscript{\rm 1} Devendra Singh Dhami,\textsuperscript{\rm 1} Kristian Kersting\textsuperscript{\rm 1, 2}
}
\affiliations{
    \textsuperscript{\rm 1}Department of Computer Science, TU Darmstadt, Darmstadt, Germany\\
    \textsuperscript{\rm 2}Centre for Cognitive Science, TU Darmstadt, and Hessian Center for AI (hessian.AI)\\
    \{yu, devendra.dhami, kersting\}@cs.tu-darmstadt.de

}

\usepackage{bibentry}

\begin{document}

\maketitle

\begin{abstract}
Probabilistic circuits (PCs) have become the de-facto standard for learning and inference in probabilistic modeling. We introduce Sum-Product-Attention Networks (SPAN), a new generative model that integrates probabilistic circuits with Transformers. SPAN uses self-attention to select the most relevant parts of a probabilistic circuit, here sum-product networks, to improve the modeling capability of the underlying sum-product network. We show that while modeling, SPAN focuses on a specific set of independent assumptions in every product layer of the sum-product network. Our empirical evaluations show that SPAN outperforms state-of-the-art probabilistic generative models on various benchmark data sets as well is an efficient generative image model.
\end{abstract}

\begin{figure*}[t]
\centering
\includegraphics[width=0.95\linewidth]{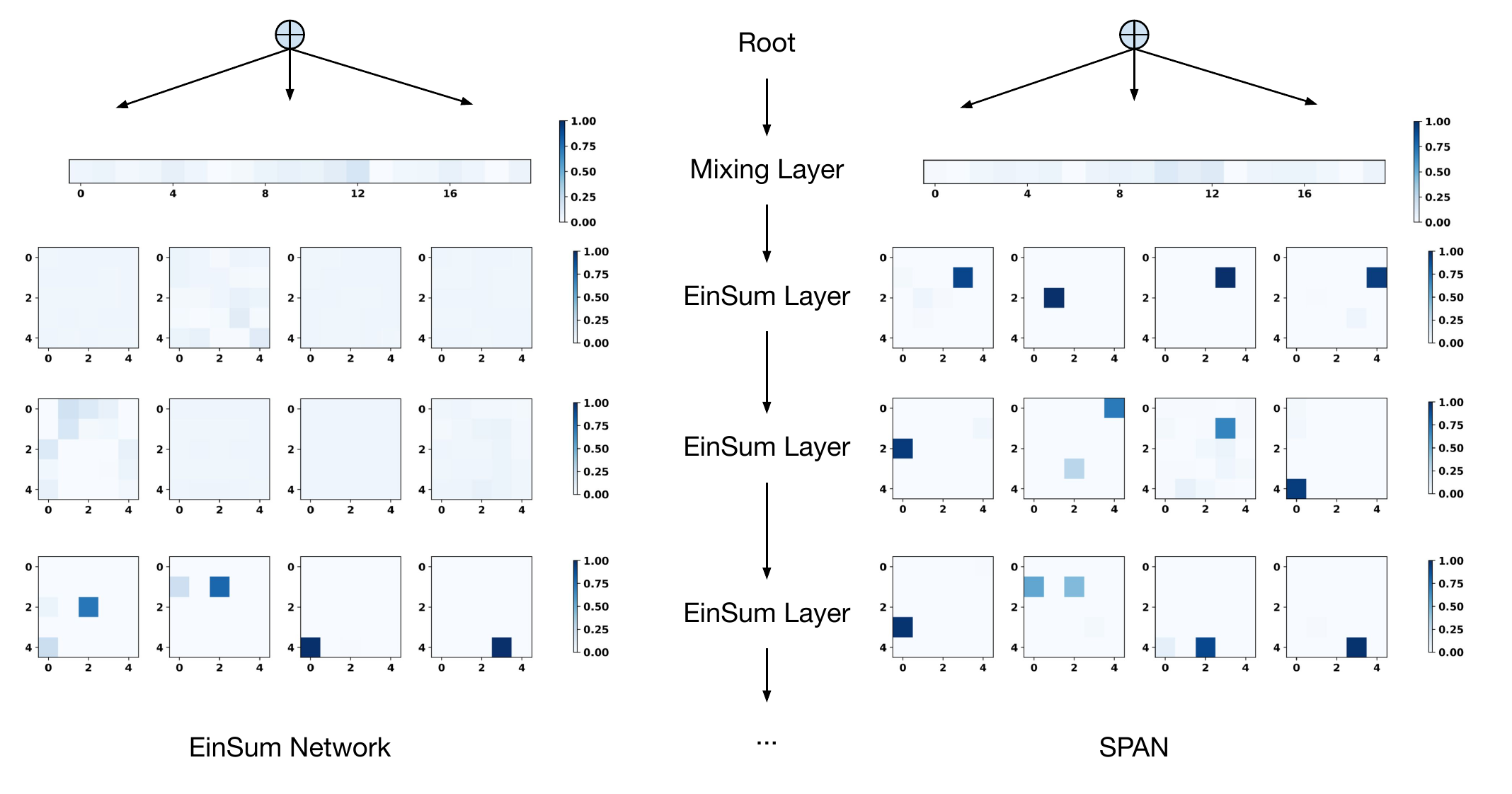} 
\caption{SPAN pays attention to only one or a few product nodes at each sum node, while EiNet treats all the product nodes from one sum node almost equally. 
The $5 \times 5$ matrices stand for the last 2 dimensions of the updated weights at a sum node, see~\cref{reweight_einet} for more details. 
The $1 \times 20$ vector visualizes the weights of the 20 replicas of the EiNet, where no self-attention is employed. Weights from one sum node are all normalized.}
\label{fig:heatmap}
\end{figure*}

\section{Introduction}
Deep learning has recently been very successful in various tasks ranging from images~\citep{pu2016variational}, videos~\citep{yue2015beyond} and text~\citep{vaswani2017attention} along with handling multi-modal data~\citep{ngiam2011multimodal,sohn2014improved}. 
Deep probabilistic models take advantage of the deep learning model efficiency and also abstract the underlying model representation while enabling reasoning over uncertainty in the domain. 
This has resulted in several probabilistic models being proposed such as arithmetic circuits~\citep{darwiche2003differential}, sum-product networks~\citep{poon2011sum} and more recently, probabilistic generating circuits~\citep{zhang2021probabilistic} to name a few. 
These probabilistic models learn the underlying joint distribution using a network polynomial over evidence variables and network parameters. 

Although probabilistic models are touted to be scalable, there are limitations when they are used for learning with real-world data. 
To alleviate these limitations, Einsum networks (EiNet)~\citep{PeharzLVS00BKG20}, which are essentially a form of sum-product networks (SPN), have been proposed. 
They combine several arithmetic operations in a single einsum operation and thus lead to learning speedups. 
Typical PCs, like EiNet, models the mixture of probability distributions at the sum node, shown in~\cref{fig:heatmap} (Left).
In the realm of deep learning, there has recently been work on Transformers which are models that leverage self-attention for capturing long-term dependencies for sequence-to-sequence modeling of text~\citep{vaswani2017attention}. 
This has given rise to very large pre-trained models like Bert~\citep{devlin2019bert} and GPT-3~\citep{brown2020language} and has had a huge impact on real-world applications of deep learning. 
There also exist models that use self-attention for density estimation~\citep{fakoor2020trade} that provide strong performance in density estimation tasks but such neural density estimators are intractable. 

In this work, we introduce Sum-Product-Attention Networks (SPAN) that incorporate the concept of self-attention in SPNs to select the most relevant sub-SPN while modeling a given data sample and also take advantage of the tractable power of probabilistic circuits. 
We show that while modeling, SPAN focuses on a specific set of independent assumptions via self-attention, in every product layer of the sum-product network. 
We specifically make use of the Transformer encoder in every product layer of SPAN. 
During the training phase of SPAN, the inputs are fed to both the Transformer encoders and the SPN. 
The encoder outputs the attention weights for the product nodes in the corresponding product layer. 
The outputs at the root of the SPN given the inputs are then dominated by the product nodes which have relatively higher attention weights, see~\cref{fig:heatmap} (Right). 

We make the following key contributions: 
(1) We introduce a new sum-product network model class that leverages the power of self-attention in order to identify and select the most relevant sub-SPN structure while learning. 
(2) Our model class is agnostic to the type of Transformer encoder as well as the SPNs being used and thus can model various data types and data distributions. 
(3) Our model offers a principled way to connect self-attention to sum-product networks, without breaking tractability in inference time. 
(4) We show that SPAN acts as an effective density estimator outperforming several state-of-the-art probabilistic circuits. 
(5) We empirically show that SPAN can learn better probability densities when compared to Einsum networks even with small structure size. 
This is expected since the use of self-attention in SPAN results in only a subset of product nodes getting activated. 

\section{Related Work}

Probabilistic circuits are popular probabilistic models which allow a wide range of exact and efficient inference routines. 
PCs were firstly developed from arithmetic circuits~\citep{darwiche2003differential}, and later on into sum-product networks~\citep{poon2011sum} and some other members such as cutset networks (CNets)~\citep{rahman2014cutset} and probabilistic sentential decision diagrams (PSDD)~\citep{kisa2014probabilistic}. 
SPNs are a family of tractable deep density estimators first presented in~\citet{poon2011sum}. 
SPNs can represent high-treewidth models~\citep{zhao2015relationship} and facilitate exact inference for a range of queries in time polynomial in the network size~\citep{poon2011sum, bekker2015tractable}. 

Several variants of SPNs have been developed to model various data distributions such as Poisson~\citep{molina2017poisson}, mixed data~\citep{molina2018mixed}, and mixture of Gaussian processes~\citep{trapp2020deep, yu2021uai_momogps}. 
SPNs can also effectively model several types of data such as relational data~\citep{nath2015learning}, graphs~\citep{zheng2018learning}, images~\citep{butz2019deep} and time series~\cite{yu2021icml_wspn}. 

The structure of an SPN encodes the dependencies of random variables at different levels. 
Hence, it is also critical to explore the structure of SPNs. 
\citet{gens2013learning} proposed to learn the structure of an SPN from data, with an iteratively splitting and clustering algorithm. 
It is also possible to randomly initialize the SPN structure and then optimize its weights, within the deep neural network fashion~\citep{peharz2020random, PeharzLVS00BKG20}, in order to achieve fast and scalable learning. 
Even a wider and deeper structure can be learned with residual links~\citep{ventola2020residual}. 
There are also special structures for specific tasks, \eg, Recurrent SPNs~\citep{kalra2018online} for sequential data, conditional SPNs~\citep{shao2020cspn} for conditional distribution and selective SPNs~\citep{peharz2014learning} to allow only one child of each sum node to have non-zero output. 
These random structures, however, do not straightforwardly unveil the dependencies among subsets of random variables. 

Transformers were initially proposed for natural language processing (NLP)~\citep{vaswani2017attention}, but have been widely adopted in various fields, such as computer vision~\citep{dosovitskiy2021image}, speech processing~\cite{dong2018speech}, databases~\citep{thorne2021natural} and genomics~\citep{ji2021dnabert}. 
 

\section{Notation and Background}

\paragraph{Notation} 
We introduce the notations used throughout the paper. 
$\mbf{X}$ is a set of random variables (RVs) with $X_i$ $\in \mbf{X}$ and 
a datum from the data set is denoted as $x$. 
$\mathcal{C}$ is a PC, where $\SumNode$ is its sum node, $\ProductNode$ is its product node, $\Leaf$ its leaf and $\Node$ a general node of a PC. 
$w_{\SumNode,\Node}$ denotes the non-negative weight of a sum node $\SumNode$ with its child $\Node$. 
$K$ denotes the number of entries of both leaf and sum layers in an EiNet. 

In a Transformer, the embedding dimension is denoted as $d_{\mathrm{model}}$, the dimension of keys is denoted as $d_k$ and the dimension of values is $d_v$. 
$h$ is the number of heads in the case of multi-head attention. 
$N \times$ denotes $N$ stacks of multi-head and feed-forward sub-layers in a Transformer encoder.

\subsection{Probabilistic Circuits}

Probabilistic circuits (PCs) are tractable probabilistic models, defined as rooted directed acyclic graphs (DAGs), in which leaf nodes represent univariate probability distributions and non-terminal nodes represent either a mixture or an independence relation of their children. 

More formally, a PC $\mathcal{C}$ over a set of RVs $\mbf{X}$ is a probabilistic model defined via a DAG $\mathcal{G}$, also called the computational graph, containing \emph{input distributions} (\emph{leaves}), \emph{sums} $\SumNode$ and \emph{products} $\ProductNode$ and a scope function $\textbf{sc}(\cdot)$. 
We refer to~\citet{PeharzLVS00BKG20} and~\citet{ TrappPGPG19} for further details. 

For a given scope function, all leaves of the PC are density functions over some subset of RVs $\mbf{U} \subset \mbf{X}$. 
This subset is called the scope of the node, and for a node $\mathsf{N}$ is denoted as $\textbf{sc}(\mathsf{N})$. 
The scope of inner nodes is defined as the union of the scope of its children. 
Inner nodes compute either a weighted sum of their children or a product of their children, \ie,~$\mathsf{S} = \sum_{\mathsf{N}\in \textbf{ch}(\mathsf{S})} w_{\mathsf{S},\mathsf{N}}\mathsf{N}$ 
and $\mathsf{P} = \prod_{\mathsf{N}\in \textbf{ch}(\mathsf{P})} \mathsf{N}$, where $\textbf{ch}(\cdot)$ denotes the children of a node. 
The sum weights $w_{\mathsf{S},\mathsf{N}}$ are assumed to be non-negative and normalized, \ie,~$w_{\mathsf{S},\mathsf{N}} \geq 0, \sum_{\mathsf{N}}w_{\mathsf{S},\mathsf{N}}=1$, without loss of generality~\citep{PeharzTPD15}. 
Further, we assume the PC to be smooth (complete) and decomposable~\citep{darwiche2003differential, choi2020pcs}. 
Specifically, a PC is \emph{smooth} if for each sum $\mathsf{S}$ it holds that $\textbf{sc}(\mathsf{N'})=\textbf{sc}(\mathsf{N''})$, for all $\mathsf{N'}, \mathsf{N''} \in \textbf{ch}(\mathsf{S})$. 
And the PC is called \emph{decomposable} if it holds for each product $\mathsf{P}$ that $\textbf{sc}(\mathsf{N'})\cap \textbf{sc}(\mathsf{N''})= \emptyset$, for all $\mathsf{N'} \neq \mathsf{N''} \in \textbf{ch}(\mathsf{P})$. 

\subsection{Transformer}

The vanilla Transformer~\citep{vaswani2017attention} is a sequence-to-sequence model consisting of an encoder and a decoder.
The encoder is mainly composed of a stack of multi-head self-attention sub-layers and position-wise feed-forward network (FFN) sub-layers.
Afterwards, a residual connection around each of the two sub-layers is employed, followed by layer normalization~\citep{he2016deep, ba2016layer}.
The decoder additionally inserts cross-attention modules between the multi-head self-attention modules and the position-wise FFNs. 
Transformer adopts attention mechanism with Query-Key-Value (QKV)
model. 
A set of queries are packed together into a matrix $Q$, and the keys and values are also packed together into matrices $K$ and $V$.
The scaled dot-product attention used by Transformer is given by:
\begin{equation}
    \mathrm{Attention}(Q, K, V) = \mathrm{softmax}(\frac{QK^{T}}{\sqrt{d_k}})V,
\end{equation}
where $d_k$ denotes the dimension of the keys, which has the same value as the queries. 

In addition to single attention function, Transformer uses multi-head attention:

\begin{align}
    \mathrm{MultiHead}(Q, K, V) &= \mathrm{Concat}(\mathrm{head_1}, \cdots, \mathrm{head_h})W^O,\\
    \mathrm{where \quad head_i} &= \mathrm{Attention}(QW_i^Q , KW_i^K , VW_i^V ),
\end{align}
where the projections are parameter matrices $W_i^Q\in \mathbb{R}^{d_{\mathrm{model}} \times d_k}$, $W_i^K \in \mathbb{R}^{d_{\mathrm{model}} \times d_k}$, $W_i^V \in \mathbb{R}^{d_{\mathrm{model}} \times d_v}$, and $W^O \in \mathbb{R}^{hd_v \times d_{\mathrm{model}}}$.

Each attention sub-layer is followed by a position-wise feed-forward network which operates separately and identically on each position:
\begin{equation}
    \mathrm{FFN}(x) = \max(0, xW_1 + b_1 )W_2 + b_2.
\end{equation}

\begin{figure*}[t]
\centering
\includegraphics[width=0.9\linewidth]{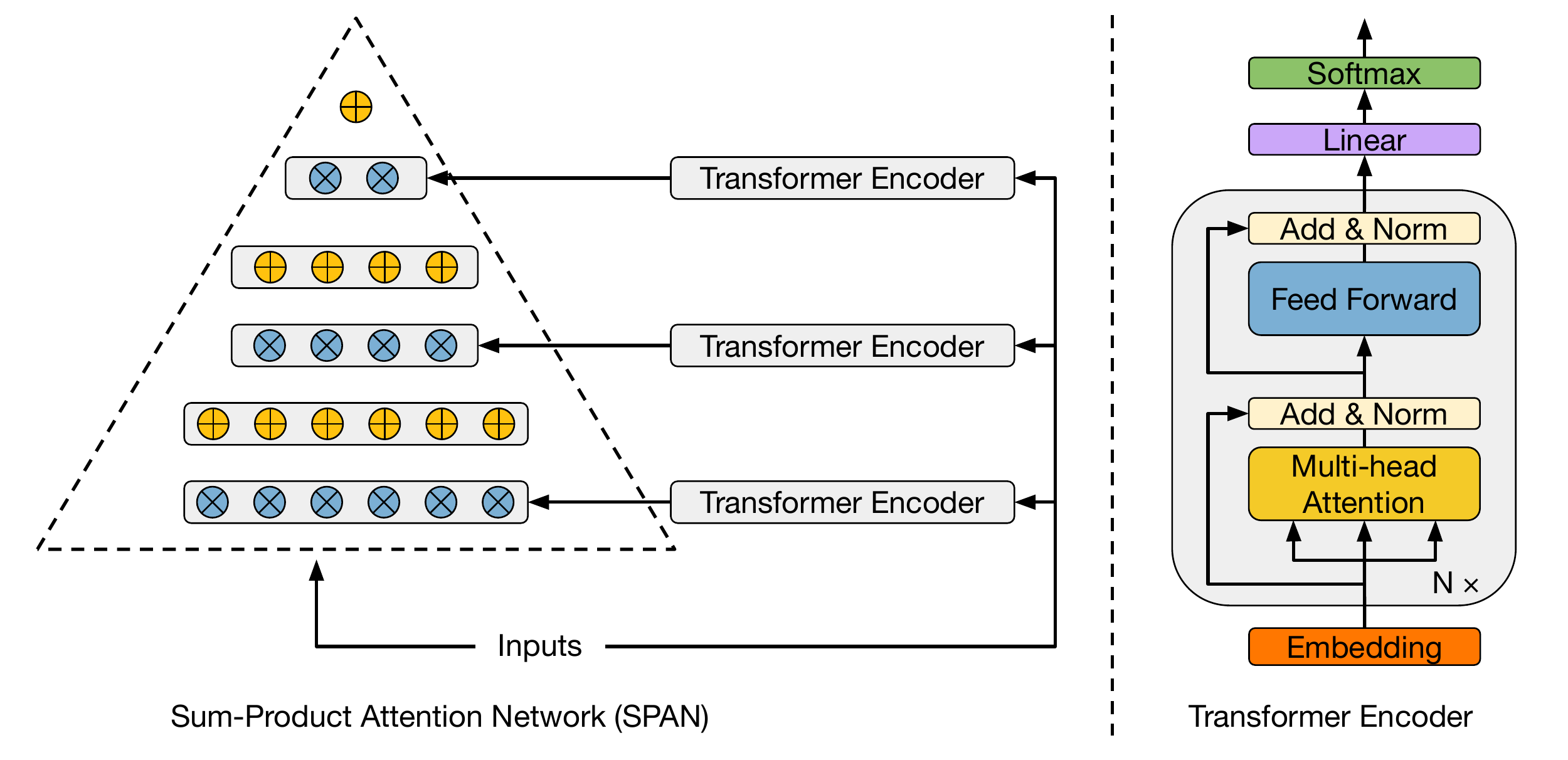} 
\caption{Illustration of Transformer encoder in SPAN. 
Inputs are fed to both the Transformer encoders and the SPN.
The Transformer encoder outputs the attention weights for the product nodes in the corresponding product layer. 
The outputs at the root of the SPN given the inputs are then dominated by the product nodes which have relatively higher attention weights.
}
\label{transformer_in_SPAN}
\end{figure*}

\section{Sum-Product-Attention Network}

The Sum-Product-Attention Network (SPAN) leverages the attention mechanism for highlighting the most relevant sub-SPNs thus improving its modeling ability. SPAN consists of an SPN and Transformers for each product layer in the SPN.
\begin{defin}
\textbf{Sum-Product-Attention Networks.}
A sum-product-attention network $\mathcal{C}^A = (\mathcal{G}, w_\mathcal{G}, \mathcal{T}, w_\mathcal{T})$ over RVs $\mbf{X} = \{X_i\}_1^N$ is an SPN $\mathcal{C} = (\mathcal{G}, w_\mathcal{G})$, connected to a set of Transformer encoders $\mathcal{T}$, 
which are parameterized by a set of graph parameters $w_\mathcal{G}$ and Transformer encoder parameters $w_\mathcal{T}$.
Given datum $x$, the normalized weights $w_{\SumNode, \ProductNode} \in w_\mathcal{G}$ of a sum node $\SumNode$ in $\mathcal{C}$ are re-parameterized by the attention weights $w_{\SumNode, \ProductNode}^A(x)$ which are outputs of $\mathcal{T}$. 
The output of $\mathcal{C}^A$ is the output computed at the root of $\mathcal{C}$.

\end{defin}

\subsection{SPAN Architecture}
Figure~\ref{transformer_in_SPAN} (Left) shows the overall SPAN architecture, where each product layer is equipped with one Transformer encoder. 
Given datum $x$, SPAN uses Transformer \emph{encoder} to produce the \textbf{attention weight} $w^{A}_{\SumNode,\ProductNode}(x)$ for the product node $\ProductNode$ where $\ProductNode \in \textbf{ch}(\SumNode)$. 
The structure of Transformer encoder in SPAN is depicted in~\cref{transformer_in_SPAN} (Right). 
The Transformer encoder consists of an embedding layer, followed by several stacks of multi-head attention sub-layers and feed-forward sub-layers. 
Similar to Transformer decoder, a feed-forward layer and a softmax layer are employed to connect the Transformer encoder output to the product nodes $\ProductNode$. 
To this end, each product node in one product layer will be provided with an attention weight from the Transformer which corresponds to this product layer. 

\subsection{Transformer Embedding}
In order to model various distributions, different embeddings can be used according to the type of random variable. 

\paragraph{Bernoulli distribution}
To model Bernoulli random variables rather than tokens as in sequence transduction models, we propose the following embedding strategy for the Transformer encoder. 
For a binary random variable $X_{i}$, value \textsf{True} is embedded as \textsf{01} and value \textsf{False} as \textsf{10}:
\begin{equation}
    \mathrm{Embedding}(X_{i}) = \begin{cases}
    \mathrm{10} & \text{ if } X_{i}= \mathrm{True}\\ 
    \mathrm{01} & \text{ if } X_{i}= \mathrm{False}
    \end{cases}.
\label{embeding_binary}
\end{equation}
Therefore, the dimension of embedding $d_{\mathrm{model}}$ for a binary random variable is $2$. 

\paragraph{Modeling images}
On the other hand, when modeling image distributions, Vision Transformer (ViT)~\citep{dosovitskiy2021image} can be employed and thus the ``Patch + Position'' embedding can be used to embed the image. 
That is, the image $X\in \mathbb{R}^{H \times W \times C}$ is reshaped into a sequence of flattened patches $X_p \in \mathbb{R}^{p \times p \times C}$, where $(H,W)$ is the resolution of the original image, $C$ is the number of (color) channels, and $p$ is the patch size. 
Here, each patch works as token and the flattened patch naturally becomes the embedding of the token. 

\subsection{Attention in SPAN}
Existing PCs use a sum node $\SumNode$ to model the mixture of its children $\Node$, \ie,~$\SumNode = \sum_{\Node \in \textbf{ch}(\SumNode)}w_{\SumNode,\Node}\Node$, where $\sum_{\Node \in \textbf{ch}(\SumNode)} w_{\SumNode,\Node} = 1$. 
In SPAN, we follow the same mechanism but pay more attention to the product nodes $\ProductNode$ that fit more the distribution of the subset of corresponding data instances. 
To achieve this, each product node $\ProductNode$ is assigned with an extra attention weight $w^{A}_{\SumNode,\ProductNode}(x)$ provided by the Transformer encoder given $x$, working as a gate. 
To be more specific, a sum node $\SumNode$ in SPAN computes the convex combination of its re-weighted children:
\begin{equation}
    \SumNode(x) = \sum_{\ProductNode \in \textbf{ch}(\SumNode)} \frac{ w_{\SumNode,\ProductNode}w^A_{\SumNode,\ProductNode}(x)\ProductNode }{\sum_{\ProductNode \in \textbf{ch}(\SumNode)} w_{\SumNode,\ProductNode}w^{A}_{\SumNode,\ProductNode}(x)},
\label{reweight_spn}
\end{equation}
where $\sum_{\ProductNode \in \textbf{ch}(\SumNode)} w_{\SumNode,\ProductNode} = 1$ and $\sum_{\ProductNode \in \textbf{ch}(\SumNode)} w_{\SumNode,\ProductNode}^A(x) = 1$. 
Thus, SPAN still models a valid probability distribution. 
Note that this operation is different from the neural CSPN~\citep{shao2020cspn}, where the weights of a sum node are directly determined by the output of the feed-forward neural network. 

SPAN can work with various types of sum-product networks. 
We illustrate here SPAN with EiNet, a novel implementation design for PCs~\citep{PeharzLVS00BKG20}, as an example. 
In EiNet, assume a sum node $\SumNode$ has one child that is a product node $\ProductNode$, and $\ProductNode$ has two children $\Node '$ and $\Node ''$. 
Therefore, the output at $\SumNode$ is given in \emph{Einstein notation} as:
\begin{equation}
    \SumNode_k = \mbf{W}_{kij}\Node '\Node'',
\end{equation}
where $\mbf{W}$ is a $K\times K\times K$ tensor, and $\Node ', \Node''$ are outputs of the children. 
$\mbf{W}$ is normalized over its last two dimensions, \ie,~$\mbf{W}_{kij} \geq 0$, $\sum_{i,j} \mbf{W}_{kij} =1 $. 
In order to provide the attention weights for the product node (here in the eimsum layer), the attention weight tensor $\mbf{W}^{A}(x)$ from the Transformer encoder should have size $b \times K\times K\times K$, given input data with batch size $b$. 
The softmax is then applied over the last two dimensions, to ensure $\sum_{i,j} \mbf{W}_{bkij}^{A}(x) =1 $. 
Following~\cref{reweight_spn}, the updated weights $\mbf{W}^{\SumNode}(x)$ would be:
\begin{equation}
    \mbf{W}_{bkij}^{\SumNode}(x) = \frac{ \mbf{W}_{kij} \times \mbf{W}_{bkij}^{A}(x)}{\sum_{i,j}\mbf{W}_{kij} \times \mbf{W}_{bkij}^{A}(x) }.
\label{reweight_einet}
\end{equation}

\subsection{Training and Inference}
The training procedure of SPAN mainly consists of 3 phases which maximize the SPAN root output representing the probability or density of the joint distribution of input data. 
In the first phase which we call SPN warm-up, the trainable parameters of the SPN are updated with $ep_1$ epochs while the Transformer parameters remain fixed. 
In the second phase, the SPN parameters and the Transformer parameters are updated iteratively, in a coordinate descent fashion for $ep_2$ epochs. 
In the last phase, the Transformer trainable parameters are again fixed and the SPN trainable parameters are fine-tuned by maximizing the root output for $ep_3$ epochs. 

The inference procedure of SPAN is similar to SPNs for taking advantage of its tractable inference property. The Transformer encoder is activated when computing the joint probability $p(\mbf{X})$. 
On the other hand, when computing the marginal distribution or the most probable explanation (MPE) from SPAN, the Transformer encoder is deactivated. 
That is, when computing the marginalization, SPAN degenerates to SPN. 
We refer to~\citet{peharz2015theoretical} for details of SPN inference. 


\begin{table}[t]
\centering
\scalebox{0.65}{
\begin{tabular}{lrrr|rcrc}
\hline
\multicolumn{1}{l}{data} & \multicolumn{1}{c}{Strudel} & \multicolumn{1}{c}{PGC} & \multicolumn{1}{c|}{RAT-SPN} & EiNet   & params  & SPAN                  & params  \\ \cline{6-6} \cline{8-8} 
                            &                             &                         &                              &         & D-R-K   & \textbf{}             & D-R-K   \\ \hline
nltcs                       & -6.06                       & -6.05                   & -6.01                        & -6.02   & 1-20-40 & {\ul \textbf{-4.35}}  & 3-20-40 \\
msnbc                       & -6.05                       & -6.06                   & -6.04                        & -6.18   & 1-\hspace{0.5em}5-40  & {\ul \textbf{-4.42}}  & 3-20-40 \\
kdd-2k                      & -2.17                       & -2.14                   & -2.13                        & -2.20   & 1-10-40 & {\ul \textbf{-2.11}}  & 2-10-40 \\
plants                      & -13.72                      & -13.52                  & {\ul -13.44}                 & -13.97  & 1-20-40 & \textbf{-13.46}       & 1-20-40 \\
jester                      & -55.30                      & -53.54                  & -52.97                       & -52.78  & 1-20-40 & {\ul \textbf{-52.07}} & 1-20-40 \\
audio                       & -42.26                      & -40.21                  & -39.96                       & -40.12  & 1-20-40 & {\ul \textbf{-39.82}} & 2-20-40 \\
netflix                     & -58.68                      & -57.42                  & -56.85                       & -56.88  & 1-20-40 & {\ul \textbf{-56.06}} & 1-20-40 \\
accidents                   & {\ul -29.46}                & -30.46                  & -35.49                       & -36.27  & 1-\hspace{0.5em}5-40  & \textbf{-35.40}       & 3-10-40 \\
retail                      & -10.90                      & {\ul -10.84}            & -10.91                       & -11.22  & 2-20-10 & \textbf{-11.06}       & 3-\hspace{0.5em}5-40  \\
pbstar                      & {\ul -25.28}                & -29.56                  & -32.53                       & -37.08  & 1-20-40 & \textbf{-36.08}       & 1-10-40 \\
dna                         & -87.10                      & {\ul -80.82}            & -97.23                       & -96.46  & 1-20-40 & \textbf{-96.10}       & 1-20-40 \\
kosarek                     & -10.98                      & {\ul -10.72}            & -10.89                       & -11.46  & 2-20-40 & \textbf{-11.40}       & 1-10-40 \\
msweb                       & -10.19                      & {\ul -9.98}             & -10.12                       & -10.53  & 1-10-20 & \textbf{-10.48}       & 2-10-40 \\
book                        & -35.77                      & {\ul -34.11}            & -34.68                       & -35.71  & 1-\hspace{0.5em}5-10  & \textbf{-35.52}       & 1-10-40 \\
movie                       & -59.47                      & -53.15                  & -53.63                       & -53.44  & 1-\hspace{0.5em}5-20  & {\ul \textbf{-53.02}} & 1-10-40 \\
web-kb                      & -160.50                     & {\ul -155.23}           & -157.53                      & -158.13 & 1-10-40 & -159.07               & 2-10-20 \\
r52                         & -92.38                      & -87.65                  & {\ul -87.37}                 & -95.93  & 1-10-40 & \textbf{-95.23}       & 2-10-40 \\
20ng                        & -160.77                     & -154.03                 & {\ul -152.06}                & -160.74 & 2-10-40 & -161.35               & 2-10-40 \\
bbc                         & -258.96                     & -254.81                 & {\ul -252.14}                & -258.49 & 2-10-40 & -260.50               & 1-\hspace{0.5em}5-40  \\
ad                          & {\ul -16.52}                & -21.65                  & -48.47                       & -46.75  & 2-10-40 & \textbf{-46.43}       & 1-10-20 \\ \hline
\end{tabular}}
\caption{SPAN outperforms EiNet on 17 of the \emph{20 binary data sets}. 
Furthermore, on 7 out of the 20 data sets, SPAN outperforms all other baselines. 
For each data set, the best performing method is {\ul underlined}, and results in \textbf{boldface} means SPAN outperforms EiNet with similar PC structures. 
The structural hyper-parameters of the EiNet show that in most cases, a larger $K$ leads to better model performance. }
\label{table:binary}
\end{table}

\section{Experimental Evaluation}

In order to investigate the benefits of SPAN compared to other probabilistic models, we aim to answer the following research questions: 
\textbf{(Q1)} Can SPAN capture probability distributions better than state-of-the-art probabilistic circuit models? 
\textbf{(Q2)} Can SPAN reduce the complexity of PC while providing comparable modeling quality? 
\textbf{(Q3)} Does SPAN work well as a generative model and out-of-distribution detector for images? 

To answer these questions, we evaluate the performance of SPAN on \emph{20 binary data sets}~\citep{lowd2010learning, haaren2012markov}, as well as the image data set \emph{SVHN}~\citep{netzer2011reading}. 
The \emph{20 binary data sets} are real-world data sets including statistics of human behaviour, text, retail, biology, etc. The number of random variables per data set ranges from 16 (``nltcs'') to 1556 (``ad''). 
These data sets are considered standard benchmarks to evaluate probabilistic circuits for density estimation~\citep{liang2017learning, dang2020strudel, zhang2021probabilistic, PeharzLVS00BKG20}. 
The \emph{SVHN} (Street View House Numbers) data set is a real-world image data set with house numbers in Google Street View images, incorporates over 600,000 digit images and comes from a significantly harder, unsolved, real-world problem~\citep{netzer2011reading}. 
\emph{SVHN} contains $32 \times 32$ RGB images of digits. 

We evaluated SPAN on the above-mentioned data sets on an NVIDIA DGX-2 with AMD EPYC 7742 64-Core Processor with 2.0TB of RAM. 
Both SPAN and EiNet experiments were executed on one NVIDIA A100-SXM4-40GB GPU with 40 GB of RAM. 
We make our code repository publicly available.\footnote{{\scriptsize \url{https://github.com/ml-research/SPAN}}}

\textbf{(Q1) Better density estimator.}
We first applied SPAN on the \emph{20 binary data sets}, to evaluate its ability of modeling Bernoulli distributions. 
The SPAN used in this section is implemented upon EiNet structure \ie,~the sum-product network component of SPAN is realized by Einsum networks. 
Similar to the hyper-parameter setting in~\citet{PeharzLVS00BKG20}, we cross-validate the split-depth $D \in \{1, 2, 3\}$, the number of replica $R \in \{5, 10, 20\}$, and the number of entries $K \in \{10, 20, 40\}$. 
The EiNet has a binary tree structure, with its weights updated by Expectation-Maximization (EM) with a step size of $0.05$. 
The Transformer encoder in SPAN, as illustrated in~\cref{transformer_in_SPAN} (Right), employs the embedding for Bernoulli distributions in~\cref{embeding_binary}, and has 2 stacks of multi-head attention sub-layer and feed-forward sub-layer. 
The Transformer weights are updated with Adam optimizer~\cite{kingma2014adam}. 

To train SPAN, we set $ep_1=2$ for warm-up training of SPN, $ep_2=5$ for coordinate descent and $ep_3=3$ for fine-tuning the SPN weights. 
We reproduced the density estimation results from EiNet, and report the results for other baselines from~\citet{dang2020strudel,zhang2021probabilistic} and~\citet{peharz2020random}. 

Table~\ref{table:binary} presents a comparison of SPAN and the baselines. On majority of the data sets, SPAN outperforms EiNet, which indicates that with the attention weights from the Transformer, the performance of the PC can be indeed improved. 
Moreover, SPAN performs best on 7 out of the 20 data sets and is comparable in the others when compared to considered probabilistic circuit baselines. 
Note that the results from PGC are from hand-crafted PGC architecture, rather than a standard along hyper-parameter search. 
Therefore, we can answer \textbf{(Q1)} affirmatively: SPAN can capture probability distributions better than the state-of-the-art PCs. 

\textbf{(Q2) Reducing the complexity of PC.} 
Another observation we can draw from~\cref{table:binary} is that, the hyper-parameter $K$ -- number of entries -- from cross-validation always results in a larger value. 
However, since SPAN is supposed to activate only a few number of the product nodes of a sum node, it is then natural to evaluate the performance of SPAN with a thinner structure. 
Therefore, we cross-validate the split-depth $D \in \{1, 2, 3\}$, the number of replica $R \in \{2, 3\}$, and the number of entries $K \in \{3, 5\}$. 
To this end, both SPAN and EiNet have a much thinner tree structure. 
Analogously, the number of children of a sum node is reduced strongly.

\begin{table}[t]
\centering
\scriptsize
\begin{tabular}{lrcrc}
\hline
\multicolumn{1}{c}{data set} & \multicolumn{1}{c}{EiNet} & params & \multicolumn{1}{c}{SPAN} & params \\\cline{3-3} \cline{5-5}
                            &                           & D-R-K  &                          & D-R-K  \\ \hline
nltcs                       & -6.08                     & 1-3-5  & \textbf{-5.19}           & 3-3-5  \\
msnbc                       & -6.27                     & 2-3-5  & \textbf{-5.17}           & 3-3-5  \\
kdd-2k                      & -2.20                     & 2-2-5  & \textbf{-2.16}           & 1-3-5  \\
plants                      & -16.08                    & 2-3-5  & \textbf{-15.73}          & 2-3-5  \\
jester                      & -53.64                    & 2-3-5  & -53.66                   & 1-3-5  \\
audio                       & -41.15                    & 1-3-5  & -41.17                   & 1-3-5  \\
netflix                     & -58.37                    & 1-3-5  & \textbf{-58.26}          & 1-3-5  \\
accidents                   & -38.21                    & 3-3-5  & \textbf{-37.81}          & 3-2-5  \\
retail                      & -11.02                    & 2-2-5  & \textbf{-10.91}          & 3-2-5  \\
pbstar                      & -36.59                    & 2-3-5  & -37.67                   & 2-3-5  \\
dna                         & -98.05                    & 3-3-5  & \textbf{-98.00}          & 3-3-5  \\
kosarek                     & -11.25                    & 2-2-5  & \textbf{-11.21}          & 2-2-5  \\
msweb                       & -10.44                    & 1-3-5  & \textbf{-10.37}          & 1-3-5  \\
book                        & -35.58                    & 2-2-5  & \textbf{-35.35}          & 2-3-5  \\
movie                       & -54.56                    & 1-3-5  & \textbf{-54.30}          & 1-3-5  \\
web-kb                      & -161.49                   & 2-3-5  & \textbf{-161.41}         & 3-3-5  \\
r52                         & -92.51                    & 3-2-5  & \textbf{-91.77}          & 3-3-5  \\
20ng                        & -159.18                   & 3-2-5  & \textbf{-158.22}         & 3-3-3  \\
bbc                         & -258.18                   & 2-3-3  & -258.71                  & 3-2-5  \\
ad                          & -56.65                    & 2-2-5  & \textbf{-46.79}          & 2-3-5  \\ \hline
\end{tabular}
\caption{With a smaller PC structure, SPAN outperforms EiNet in 16 out of the \emph{20 binary data sets}. 
SPAN even outperforms all the other baselines on data sets ``nltcs'' and ``msnbc'', with such small PC structure.}
\label{table:small}
\end{table}

We can see from~\cref{table:small} that SPAN outperforms EiNet in 16 out of 20 data sets. 
This means SPAN can improve the performance of PCs which have relatively small structure sizes. 
Nevertheless, even with such thin structures, SPAN outperforms the baselines on data sets ``nltcs'' and ``msnbc'', while the corresponding EiNet performance can not compete with the baselines. 

Additionally, we visualize this feature of SPAN qualitatively in~\cref{fig:heatmap}. 
Both EiNet and SPAN in~\cref{fig:heatmap} have identical hyper-parameters, with three einsum layers and one mixing layer on top. 
The number of entries $K$ was set to 5, the split-depth $D$ to 3, and the number of replica $R$ was 20. 
Following~\citet{PeharzLVS00BKG20}, the EiNet and SPAN are both trained on one of the 100 clusters by k-means from \emph{SVHN} data set. 
The EiNet is trained with EM for 10 epochs and SPAN training follows the 3-phase training as before, again with $ep_1=2$ for warm-up, $ep_2=5$ for coordinate descent and $ep_3=3$ for fine-tuning. 

As shown in~\cref{fig:heatmap} (Left), the weights of the EiNet, especially in the upper layers, are mostly equally distributed. 
That means, all the product nodes of a sum node contribute to model the mixture of probability distributions at the sum node. 
However, in SPAN, only one or a few product nodes dominate the mixture of a sum node, visualized with dark-blue entries in~\cref{fig:heatmap} (Right). 
Therefore, SPAN activates only a small subset of the product nodes at a sum node. 
Note that the weights in EiNet are fixed once it is trained, while the weights in SPAN are influenced by the attention weights $w^{A}_{\SumNode,\ProductNode}(x)$ from the Transformer encoder output, given input $x$. 
Hence, the activated product nodes are determined by the Transformer encoder given input data showing the importance and effectiveness of attention weights in SPAN. 

In the inference routine, most of the product nodes actually contribute negligibly to their parents. 
With only one or a few product nodes activated, SPAN can already provide better or at least equivalent modeling power. 
Thus, \textbf{(Q2)} can be answered affirmatively: SPAN can provide comparable modeling quality with a much smaller structure of its PC.

\begin{figure}[t]
\centering
\includegraphics[width=0.9\columnwidth]{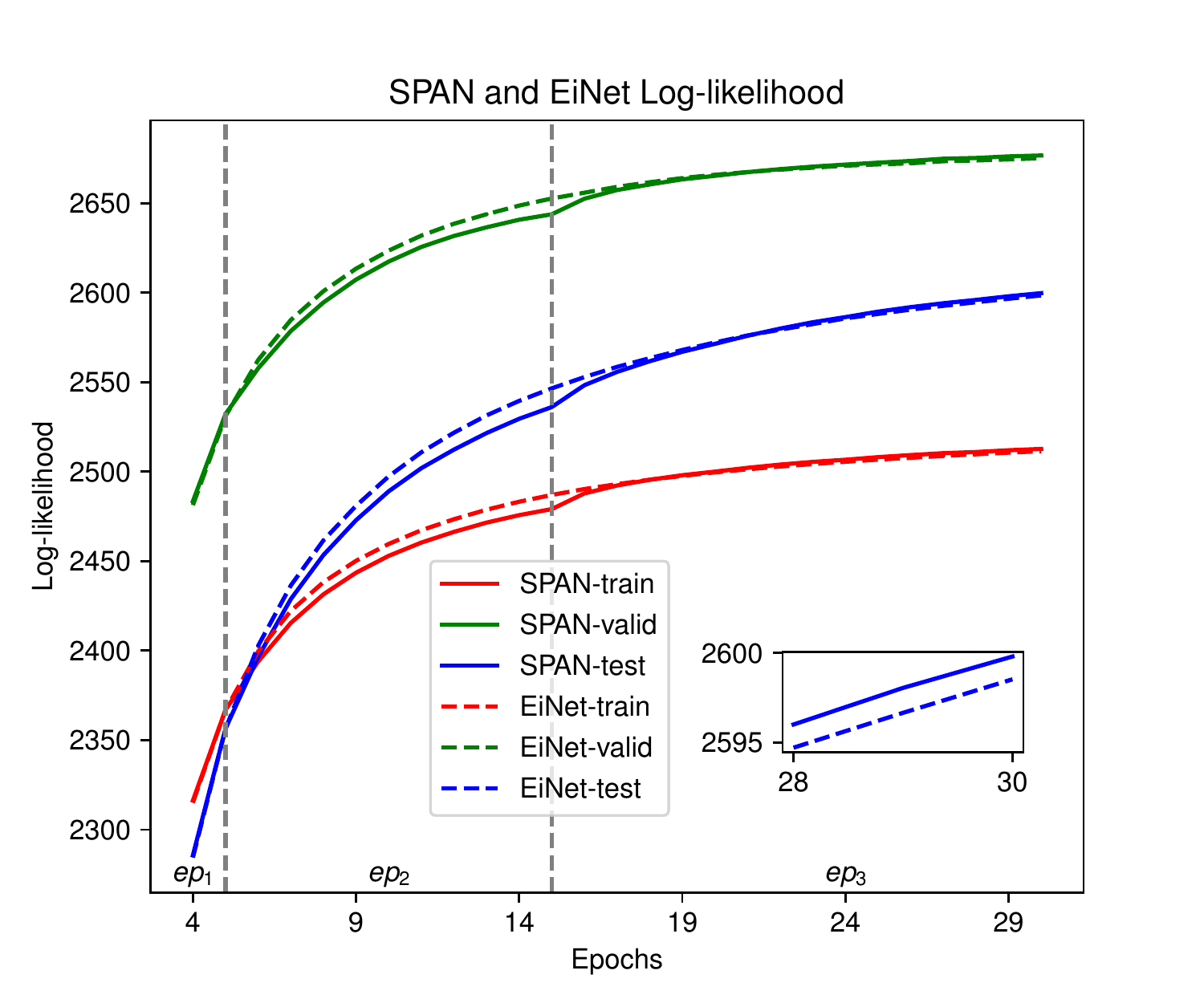} 
\caption{Training, validation and test log-likelihoods of SPAN and EiNet, on \emph{SVHN}. 
The convergence of SPAN is a bit slower than EiNet during the coordinate descent phase, but SPAN converges faster than EiNet in the fine-tuning phase and obtains a higher log-likelihood. 
Final log-likelihoods of the test set are visualized in the zoom-in plot. }
\label{fig:svhn_loss}
\vspace{-0.2in}
\end{figure}

\begin{figure}[h!]
\centering
\includegraphics[width=0.95\columnwidth]{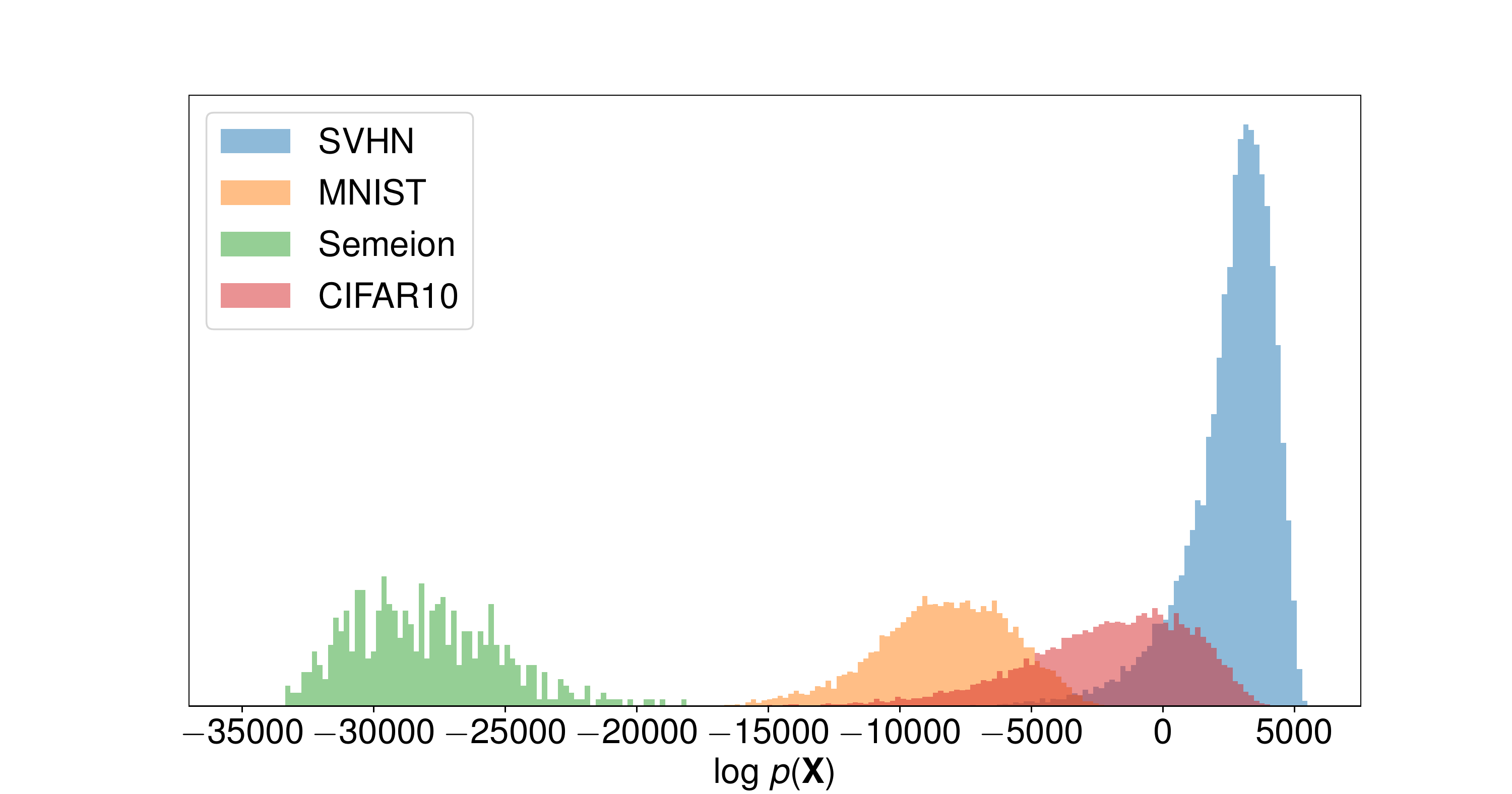} 
\caption{Sample-wise SPAN log-likelihoods of \emph{SVHN} test set are much higher than the test sets of  MNIST, Semeion and CIFAR10. 
SPAN is trained on \emph{SVHN} training set. 
Bins of Semeion data set are re-scaled for better visualization.}
\label{fig:hist}
\end{figure}

\textbf{(Q3) Generative model for images and Out-of-distribution detection.} 
We train SPAN on \emph{SVHN} data set to evaluate its modeling ability for images. 
The EiNet has number of entries $K=5$, split-depth $D=4$, and number of replica $R=100$, with a random binary tree structure. SPAN employs the same EiNet structure, and uses a Vision Transformer to encode the image inputs. 
The ViT has patch size 8, depth 2, and 16 heads and the embedding has dimension 1024 with feed-forward layers of dimension 512.
The EiNet and SPAN are both trained on the full \emph{SVHN} data set. 
While EiNet is trained with EM for 30 epochs, SPAN training follows the 3-phased training as before with number of epochs in each phase increased to to $ep_1=5$ for warm-up training of SPN, $ep_2=10$ for coordinate descent and $ep_3=15$ for fine-tuning the SPN weights with fixed Transformer weights. 
The step size of EM is $2\mathrm{e}{-2}$ and the learning rate of Adam for updating the Transformer is $1\mathrm{e}{-7}$.

\begin{figure*}[t]
\centering
\begin{minipage}[t]{0.24\linewidth}
  \centering
  \vspace{-5.8cm}
  \includegraphics[width=0.9\textwidth]{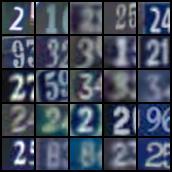}\\
  (a) Real \emph{SVHN} Images
\end{minipage}
\begin{minipage}[b]{0.24\linewidth}
  \centering
  \includegraphics[width=0.9\textwidth]{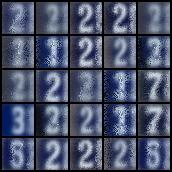}\\
  (b) SPAN Samples\\
  \includegraphics[width=0.9\textwidth]{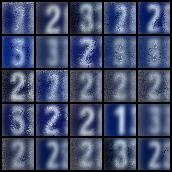}\\
  (c) EiNet Samples
\end{minipage}
\begin{minipage}[b]{0.48\linewidth}
  \centering
  \includegraphics[width=0.9\textwidth]{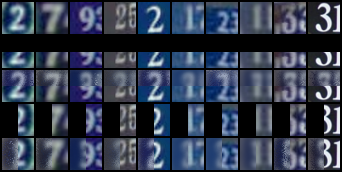}\\
  (d) SPAN Reconstructions \\
  \includegraphics[width=0.9\textwidth]{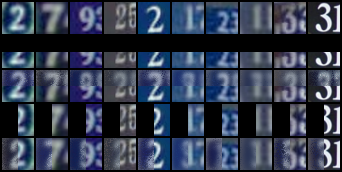}\\
  (e) EiNet Reconstructions
\end{minipage}
\caption{SPAN generates samples as good as EiNet and reconstructs better visualized digits. }
\label{fig:svhn}
\end{figure*}

The training, validation and test log-likelihoods from both SPAN and EiNet are plotted in~\cref{fig:svhn_loss}. 
Both SPAN and EiNet perform similarly during the first 5-epoch EM updates. 
When switching to coordinate descent, the log-likelihood of SPAN increases a bit slower than that of EiNet, as the PC in SPAN has to adapt to the newly updated attention weights. 
However, when switching back to EM update for the PC, the log-likelihood of SPAN quickly raises. 
In the end, the average test log-likelihood from SPAN ($\mbf{2599.80}$) reaches even higher than the EiNet ($\mbf{2598.53}$). 
In other words, SPAN does not undermine the performance of its PC, while still being able to activate the most relevant children of sum nodes. 
We ran both SPAN and EiNet for 5 times, and the average running time for SPAN is 11341.9s and for EiNet is 5547.5s. 

SPAN can also detect the out-of-distribution test samples which is a property of PCs. 
As shown in~\cref{fig:hist}, the log-likelihoods of MNIST~\citep{lecun1998gradient} test set have almost no overlap with the \emph{SVHN} log-likelihoods, mainly because MNIST has grayscale images. 
CIFAR10~\citep{krizhevsky2009learning} has color images thus distributes closer to \emph{SVHN}, while the overlap happens at the lower log-likelihood values of \emph{SVHN}. 
The Semeion~\citep{buscema1998metanet} data set contains binary images thus has extremely low log-likelihoods, with an average of $-28065$. 
Hence, SPAN trained on \emph{SVHN} can successfully distinguish the outliers from MNIST and Semeion data sets, and also provide much lower log-likelihood given images from CIFAR10.

To qualitatively evaluate SPAN as generative image models, we visualize the samples and reconstructions from SPAN in~\cref{fig:svhn}. 
Similar to the experimental settings in \textbf{(Q2)}, both SPAN and EiNet are trained on one cluster from 100 k-means clusters from \emph{SVHN} data set. 
Figure \ref{fig:svhn} (b) and (c) show samples from SPAN and EiNet, respectively. 
SPAN produces image samples as good as EiNet. 
There is no ``stripy'' artifacts as both SPAN and EiNet employed a binary tree structure, instead of the PD architecture~\citep{poon2011sum, PeharzLVS00BKG20}. 
On the other hand, the sampled images contain more noise. 
The blurry samples are due to the blurry images in the data set, see~\cref{fig:svhn} (a). 

In most cases, both SPAN and EiNet successfully reconstructed the digit given half of the image, as shown in~\cref{fig:svhn} (d) and (e). 
Overall, the SPAN reconstructions show less noise than EiNet, \eg, digit ``7'' in the $2^{nd}$ column, digit ``9'' in the $3^{rd}$ column, and digit ``2'' in $4^{th}$ and $5^{th}$ columns. 
Furthermore, SPAN better reconstructs digit ``2'' in the $7^{th}$ column, especially in the case of left-half-missing. 
The reconstruction from SPAN is a left half of digit ``2'', while the EiNet reconstruction is not recognisable. 
The reconstruction of this image is challenging as there is also another digit ``3'' in the image, which inputs additional noise to the models. 

As a conclusion, SPAN can model the distribution of images well, providing samples as good as baseline, and producing better visualized reconstructions than the considered baseline. 
Therefore, \textbf{(Q3)} can be answered affirmatively.

\section{Conclusion}
We presented SPAN, a new model that incorporates attention in the sum-product network architecture. This results in selection of the most relevant sub-SPN structures during learning while taking advantage of the tractable power of probabilistic circuits. We show that SPAN is a better density estimator than several state-of-the-art probabilistic models and can also learn well given a thinner SPN structure. Future works include reducing the number of embeddings while modeling Bernoulli random variables. Extending our model to handle missing data and marginalization with the Transformer encoder is a natural next step. 

\section*{Acknowledgements}
This work was supported by the Federal Ministry of Education and Research (BMBF; project ``MADESI'', FKZ 01IS18043B, and Competence Center for AI and Labour; ``kompAKI'', FKZ 02L19C150), the ICT-48 Network of AI Research Excellence Center ``TAILOR'' (EU Horizon 2020, GA No 952215), the Collaboration Lab ``AI in Construction'' (AICO), the Hessian Ministry of Higher Education, Research, Science and the Arts (HMWK;
projects ``The Third Wave of AI'' and ``The Adaptive Mind''), and the
Hessian research priority programme LOEWE within the
project ``WhiteBox''.




\bibliography{zhongjie.bib, aaai22}

\end{document}